\documentclass[runningheads]{llncs}
\usepackage[T1]{fontenc}
\usepackage{graphicx}
\usepackage{amsmath}
\usepackage{rotating}  
\usepackage{booktabs}
\usepackage{multirow}
\usepackage{booktabs} 
\usepackage{array}

\newcolumntype{P}[1]{>{\raggedright\arraybackslash}p{#1}}

\begin{document}
%

\title{Anatomy-Aware Promptable Segmentation with Online Interactive Training for AUTOPET V}

\titlerunning{AUTOPET V challenge}
%
\author{Pablo Lozano-Jimenez\inst{1,2}\orcidID{0009-0001-1654-4841} \and
Sergio Romero-Tapiador\inst{2}\orcidID{0000-0002-8919-8687} \and
Ruben Tolosana\inst{2}\orcidID{0000-0002-9393-3066}}
\authorrunning{P. Lozano Jimenez et al.}
%
\institute{University of Amsterdam, Faculty of Science, Amsterdam, The Netherlands \and
BiometricsAI, Universidad Autónoma de Madrid, 28049 Madrid, Spain
}
\maketitle              
\begin{abstract}
We present an anatomy-aware, promptable model for whole-body lesion segmentation in FDG and PSMA PET/CT, developed for the AUTOPET V challenge. The proposed method is built as family of nnU-Net-based models and trained in two stages: \textit{i)} a pre-training stage that produces a strong initial segmentation, and \textit{ii)} an online interactive stage that learns to exploit scribble prompts, refining the prediction over successive interactions. Anatomical context is incorporated through organ supervision using a single shared head that predicts lesions and organs from the same features, which reduces false positives arising from physiological uptake. Also as the tracer (i.e., FDG/PSMA) is not provided at inference, we add a tracer classifier based on image processing and a random forest over coronal MIP features, routing each study to a combined FDG+PSMA model or to a PSMA-specific model. Across four-fold cross-validation the organ-supervised model achieves the best and most stable performance, the interactive stage improves the Dice score monotonically with each prompt, and PSMA-specific training yields the strongest tracer-wise results.

\end{abstract}
\section{Introduction}
PET and CT studies play an important role in the detection and diagnosis of cancer. In current practice, radiologists manually assess tumors by visually looking at the lesions and changes in size and anatomy around them. AI models have been recently developed to automatize this process, addressing manual segmentation problems, such as inter-observer variability  and the difficulty to detect complex data patterns. However, state-of-the-art models still struggle with domain shifts \cite{doadap}, low-contrast lesions, and diverge from the clinical perspective \cite{standards}.\\

The AUTOPET V challenge addresses these gaps by analyzing interactive human-AI segmentation which makes the cancer detection faster, more accurate and closer to the clinical practice. The challenge is built on a large and heterogeneous public dataset comprising 1,014 FDG studies \cite{dataapV} \cite{dataapV1} and 597 PSMA studies \cite{dataapV2}, providing the variability required to test model generalizability and to ensure the validity of the reported results.\\

The goal of the current edition is to improve anomaly-detection performance by introducing a human-in-the-loop scheme that closely mimics clinical interaction. In other words, the aim of this challenge is to obtain accurate lesion segmentation, correctly delineating the most difficult lesions with the least possible interaction effort. To this end, the number of allowed interactions is reduced with respect to last year's edition (from 10 to 5), and a new form of clinical validation is introduced in which the test cases contain interactions performed by specialist physicians who mark the lesions that are hardest to identify.\\ 

The proposed method is based on the latest version of the nnU-Net \cite{nnunet} framework, currently one of the most influential frameworks in medical image segmentation. It delivers strong results and provides the preprocessing needed to combine the PET and CT modalities. We further incorporate ideas from previous AUTOPET editions \cite{apIV1}\cite{apIV2}\cite{apIII}  and related scientific literature; these were compared and merged with our own ideas in order to find the most effective and optimal model for the task. Our proposed METHOD is publicly available in GitHub: \url{https://github.com/BiometricsAI/AUTOPET\_V\_submission}

\section{Methods}

\subsection{Data}
\subsubsection{Training data.}
The AUTOPET V public dataset was used for training and validation. The data combine PET/CT scans performed using different tracers (FDG and PSMA). There are 1,014 studies involving 501 cases with cancerous lesions, including lung cancer, lymphoma and melanoma and 513 cases with negative controls using FDG.  Regarding the PSMA studies, 537  involved patients and 60 were negative control studies. All these studies were conducted on patients with varying age and gender distributions. The dataset is therefore balanced and free from significant bias.  The FDG-tracer cases were all acquired at a single center with the same scanner (Siemens Biograph mCT PET/CT). In contrast, the PSMA-tracer cases were acquired with three different scanners (Siemens Biograph 64-4R TruePoint, Siemens Biograph mCT Flow 20, and GE Discovery 690). This acquisition heterogeneity introduces variability in the imaging protocol and in the quality and resolution of the studies, and highlights the need for methods that address the resulting domain gap.

\subsubsection{Validation data.}
To obtain a robust validation strategy, we adopted a cross-validation scheme in which the full AUTOPET V dataset was split into 4 folds. So that the folds are balanced and the experimental protocol is well defined, each fold contains the same number of cases per tracer (FDG and PSMA) and, in particular, per category. Concretely, we performed a class-stratified split, in which for FDG the categories are negative, melanoma, lymphoma, and lung cancer. For PSMA, on the other hand, the categories are negative and each of the three scanners.

\subsection{Data pre-processing}
Preprocessing followed the standard nnU-Net protocol \cite{nnunet}. Intensity normalization is applied per channel according to its modality. For the CT channel, images were normalized with the "CT normalization" is applied, while for the PET channel, per-image z-score normalization is applied. Finally, the network topology, patch size, and batch size are configured automatically from the dataset fingerprint (voxel spacings, image sizes, and intensity distributions).

\subsection{Proposed method}
To define our proposed model we followed an incremental methodology, progressively incorporating improvements in order to ultimately assemble a complete final system. So that the model can produce a good initial segmentation and subsequently learn in a meaningful, positive way from scribble interactions, training is carried out in two phases:
\begin{itemize}
    \item Phase-1 (pre-training): two input channels are provided to the network: the PET and CT studies. This phase enables the model to produce an accurate initial segmentation.
    \item Phase-2 (interactive learning): prompts are added incrementally at each iteration, and the model learns online from its own errors at every step.
\end{itemize}

Our method is built as an incremental family of models, where each one extends the previous with a single, well-motivated change, so the contribution of every design choice can be evaluated in isolation. All are trained as Phase-1 backbones, and the best one becomes the pre-trained starting point for the Phase-2 interactive stage.\\

Regarding Phase-1 (pre-training), starting from a baseline nnU-Net that stacks PET and CT as input channels (Version 1), Version 2 adopts a modality-aware design with a Siamese encoder and separate PET, CT, and joint decoders; Version 3 adds organ supervision through a shared lesion–organ head; and Version 4 fuses both, using two independent modality-specific encoders feeding three task-specialized decoders (organ segmentation from CT, lesion detection from PET, and refined lesion segmentation from the fused features). Finally, in Version 5, at the system level, we account for the fact that the acquisition tracer is not provided at inference: a lightweight tracer classifier routes each study either to the model trained on the combined FDG+PSMA data or to a variant of the same model trained exclusively on PSMA, following evidence that tracer-specific training benefits PSMA cases. We provide next further details for all the versions.

\paragraph{\textbf{Phase-1 - Version 2. Architecture.}} We first implemented an architecture that departs from the original nnU-Net. Instead of a single encoder and a single decoder, we used two Siamese (weight-sharing) encoders that encode the PET and CT studies respectively, followed by three decoders: a PET decoder, a CT decoder, and a joint decoder that operates on the combined information obtained by concatenating the encoder features at each resolution level (before the skip connections). This architecture has already been successful for the detection of lymphomas \cite{lymph}.
This design allows the model to capture anatomical and functional information separately through the respective decoders, forcing it to segment correctly in each modality independently as well as from the combined information of both.
The model is trained with a combined loss defined as
\begin{equation}
    L = 1.0 \cdot L_{\text{joint}} + 0.3 \cdot L_{\text{PET}} + 0.3 \cdot L_{\text{CT}}
\end{equation}

where each term is the standard nnU-Net DICE + Cross-Entropy loss.

\paragraph{\textbf{Phase-1 - Version 3. Organ supervision.}} Following the success reported in previous AUTOPET editions \cite{apIV1}\cite{apIV2}\cite{apIII}, our  proposed model also incorporates organ supervision, although in a form that initially differs from prior work. Rather than using two separate segmentation heads that perform lesion and organ segmentation independently and are merged during training through a joint loss, our Version 3 uses a single encoder and a single decoder for the joint segmentation of both structures. The final convolution of the decoder is widened so that it produces both the lesion logits and the organ logits from the same feature map, with two independent softmax (a voxel may be, e.g., both a lesion and part of the liver). The lesion and organ losses are weighted equally.\\

This coupling is the core of the design. Because the lesion logits are produced from the very same feature vector that must also predict organ identity, the network is forced to encode anatomical context directly into the representation used for lesion segmentation, instead of confining it to a separate branch that it could otherwise learn in isolation. In effect, this acts as a strong anatomical regularizer. The model learns where physiological uptake is expected and stops reporting it as lesion, which markedly reduces false positives. Equal weighting of the two tasks (as in article \cite{apIV1}) is what makes this regularization effective and down-weighting the organ task leaves it essentially inert. \\

To build this model, we first augmented the ground truth with segmentation labels for 9 organs: spleen, kidneys, liver, urinary bladder, lung, brain, heart, stomach, and prostate. They were obtained with TotalSegmentator model \cite{totalseg}.

\paragraph{\textbf{Phase-1 - Version 4. Combined model: detection + anatomical context.}} 
In this version we combine the two improvements proposed before. We therefore define a model with two separate encoders with independent weights that encode the PET and CT images separately. The outputs of both encoders are then combined in three decoders that perform independent tasks: the CT decoder, which performs organ segmentation using the labels produced by TotalSegmentator; the PET decoder, which performs lesion detection; and the joint PET+CT decoder, which fuses the latent features of both encoders and performs the actual cancer-lesion segmentation.\\

Through this task guidance and separation, each decoder learns a specific and independent task.

\paragraph{\textbf{Phase-1 - Version 5. Tracer classifier and PSMA-specific training.}} Previous approeaches (\cite{apIV1} and \cite{apIII})  reported improved performance when training a separate model on the PSMA data. Following this evidence, we trained a dedicated model using the same architecture as our Version 3, but on PSMA studies only.\\

As the inference data do not indicate which tracer (FDG/PSMA) each study was acquired with, we implemented a tracer classifier. While this idea has already been used by previous AUTOPET winners, our model differs. We observed that a deep classifier trained on AUTOPET V dataset generalized poorly, achieving low accuracy when tested on an external dataset such as the one developed for DeepPSMA challenge\cite{deeppsmadata}. We therefore implemented a more generalizable version based on image processing and classical machine learning, designed to exploit the differences in the natural physiological uptake of each tracer.\\

The pipeline first computes coronal maximum-intensity projections (MIPs) of the PET studies. The projection is then partitioned into three regions: head, abdomen, and pelvis; and the intensity of each region is analyzed relative to the total image intensity. The discriminative differences are that, with FDG, the brain shows high uptake driven by glucose metabolism and the abdomen exhibits a more irregular uptake pattern, whereas PSMA naturally concentrates in the abdomen and in the cranial (salivary and lacrimal) glands. To determine whether cranial uptake originates from the brain or from these glands, we perform a topological analysis of the intensity profile along the transverse axis, extracting features such as FWHM ratio, FWHM85 ratio, the number, spacing and asymmetry of peaks, and the valley depth. These features, together with the intensity proportion in each ROI, are used to train a random forest using AUTOPET V dataset that serves as the classifier, achieving an accuracy of 98\%.\\

To guard against misrouting, a study is assigned to the PSMA-specific model only when the classifier's confidence exceeds 65\%. Uncertain cases are segmented with the model trained on the combined dataset. Because the combined model has seen both tracers, it is the safe fallback for ambiguous studies, so classification errors on cases degrade gracefully rather than sending an FDG study to a PSMA-only model. By applying this 65\% confidence threshold, all cases predicted as PSMA are classified with 100\% accuracy.

\paragraph{\textbf{Phase-2 - Interactive simulations.}} 
The interactive stage (Phase-2) is not trained from scratch: it is obtained by fine-tuning the Phase-1 version 3 and 5 model, so that at the first interaction step the network reproduces the Phase-1 segmentation and then learns to exploit the scribbles from that starting point. Two Phase-2 checkpoints are produced, one per tracer route. For FDG studies, we fine-tune the fold that performed best for this tracer among the models trained on the combined FDG+PSMA dataset (version 3). For PSMA studies, we instead fine-tune the best-performing fold obtained from training on the independent PSMA-only dataset (version 5).

To incorporate prompts in the human-in-the-loop setup, the input scribbles were transformed into Gaussian heatmaps and added as supplementary foreground and background channels. Furthermore, to evaluate the influence of the Gaussian standard deviation, we assessed two different prompt encoding configurations using gaussian standard deviation ($\sigma$) values of 5 and 10.\\

In addition, following the success of \cite{apIV1}, we adopted a stochastic prompt-sampling policy. To prevent the model from becoming \textit{lazy}, always learning to segment under the same fixed number of interactions, and to better mimic clinical interaction, the number of iterations (i.e., simulations) per training case is not deterministic but is drawn from a distribution with the following probabilities: [0.24,0.18,0.13,0.09,0.11,0.25] (the number of simulated segmentation steps equals the number of scribbles plus one). By assigning a high probability to the 0-scribble case we safeguard good initial segmentation and ensure that the system does not forget the pre-training obtained in Phase-1.\\

Finally, the paradigm used in this system is online training. Rather than injecting all prompts statically from the beginning of training, they are added progressively. For each case, training starts with no interaction. This initial segmentation is then added as an additional channel. Then, a scribble is simulated on the lesion with the largest error region and encoded as an additional foreground/background channel, and finally a new segmentation is produced. The updated segmentation mask and the new prompt are fed as additional channels for the next interaction, and the process is repeated for the number of iterations defined for each case. 

The Phase-2 model is trained with a compound loss combining three terms: the lesion loss, an auxiliary organ loss, and a scribble-consistency loss. The lesion and organ terms are both Dice + cross-entropy. The scribble term makes the prediction agree with the prompts at the marked voxels. The total loss is
\begin{equation}
L = \lambda_{les}\, L_{les} + \lambda_{org}\, L_{org} + \lambda_{scr}\, L_{scr},
\end{equation}
with $\lambda_{les}=1.0$, $\lambda_{org}=0.5$ and $\lambda_{scr}=0.5$. These values were selected experimentally.

\subsection{Data post-processing}
No data post-processing has been applied for this model.

\subsection{Training and test parameters}
All models were trained within the nnU-Net v2 \cite{nnunet}  framework, which automatically configures the network topology, patch size, and normalization from the dataset fingerprint. Phase-1 (pre-training) was trained with SGD (momentum 0.99, Nesterov) under nnU-Net's default polynomial-decay learning-rate schedule with an initial learning rate of $10^{-2}$, a batch size of 2, and deep supervision; the loss was Dice + Cross-Entropy for the lesion head and, for the organ-supervised model (Version 3), an additional equally-weighted Dice + Cross-Entropy term on the organ head (with an ignore label for cases lacking organ annotations), both produced by a single shared, widened final convolution. Phase-2 (online interactive stage) was obtained by fine-tuning the best Phase-1 fold: the network was warm-started from the Phase-1 weights, the guidance-injection convolutions were zero-initialized so that the model initially reproduces the Phase-1 segmentation, and training used SGD with warm-up (10 epochs) followed by polynomial decay, an initial learning rate of $10^{-3}$ with a reduced (0.1x) rate on the encoder, gradient-norm clipping, and 500 epochs. At each step a scribble is simulated on the largest error region and encoded as an additional foreground/background guidance channel via a Gaussian heatmap. The number of interactions per case (0–5) is drawn from a stochastic distribution with elevated probability at zero to preserve the initial segmentation, and the total loss combines the lesion Dice + CE, the auxiliary organ term, and a scribble-consistency term.

\section{Results}
\paragraph{\textbf{Phase-1.}} Table~\ref{phase1} reports the four-fold cross-validation results of the successive model versions. Moving from the baseline (Version 1) to the modality-aware architecture (Version 2) already improves the final metric but the largest single gain comes from adding organ supervision in Version 3. The improvement of Version 3 over Version 2 is driven by anatomical context. By forcing the network to predict organ identity from the same features that produce the lesion logits, the model learns where physiological uptake is expected and stops reporting it as disease. The effect is clearest on FDG, where physiological glucose uptake in healthy organs is the dominant source of false positives. PSMA, however, due to its more stereotyped biodistribution, benefits less. Interestingly, the combined dual-encoder model (Version 4) does not improve over Version 3. Its overall metric is essentially identical to Version 3, indicating that the additional architectural complexity of two specialized encoders and task-separated decoders provides no measurable benefit here.\\

Version 3 was therefore selected as the model to carry into Phase~2, for two reasons. First, it attains the best final metric. Second, and decisively, it exhibits a substantially lower cross-fold standard deviation than Version 4. Low variance is particularly important for Phase~2. As the interactive stage is fine-tuned from a single fold (see below), a low-variance backbone means the folds are interchangeable and the chosen checkpoint is representative of true performance rather than a fortunate or unfortunate draw. A high-variance model like Version 4 risks starting the interactive stage from an outlier fold. Finally, Version 5 introduces the tracer classifier together with PSMA-specific training and yields the best results of all versions. On PSMA it reaches a result which is clearly above the one obtained by Version 3 trained on the full combined FDG+PSMA dataset, confirming that a PSMA-dedicated model outperforms the generalist on PSMA studies.

\begin{table}[htbp]
\centering
\caption{Phase-1 results. Training and evaluation were done using AUTOPET V Dataset. V1: Baseline; V2: Architecture; V3: Organ supervision;
V4: Combined model; V5: Tracer classifier \& PSMA-specific training}
\label{phase1}
\renewcommand{\arraystretch}{1.3}
\setlength{\tabcolsep}{6pt}
\begin{tabular}{ll ccccc}
\toprule
 & & \textbf{V1} & \textbf{V2} & \textbf{V3} & \textbf{V4} & \textbf{V5} \\
\midrule
\multirow{3}{*}{\textbf{Overall}}
 & DICE   & $0.449{\pm}0.024$ & $0.507{\pm}0.017$ & $0.551{\pm}0.025$ & $0.557{\pm}0.067$ & $\mathbf{0.578{\pm}0.019}$ \\
 & DMM    & $0.752{\pm}0.013$ & $0.767{\pm}0.014$ & $0.785{\pm}0.019$ & $0.776{\pm}0.019$ & $\mathbf{0.794{\pm}0.020}$ \\
 & Metric & $0.600{\pm}0.019$ & $0.637{\pm}0.015$ & $0.668{\pm}0.022$ & $0.666{\pm}0.043$ & $\mathbf{0.686{\pm}0.020}$ \\
\midrule
\multirow{3}{*}{\textbf{FDG}}
 & DICE   & $0.415{\pm}0.028$ & $0.498{\pm}0.018$ & $0.559{\pm}0.032$ & $\mathbf{0.573{\pm}0.093}$ & $-$ \\
 & DMM    & $0.606{\pm}0.028$ & $0.629{\pm}0.009$ & $\mathbf{0.670{\pm}0.025}$ & $0.654{\pm}0.029$ & $-$ \\
 & Metric & $0.511{\pm}0.028$ & $0.563{\pm}0.014$ & $\mathbf{0.614{\pm}0.028}$ & $0.613{\pm}0.061$ & $-$ \\
\midrule
\multirow{3}{*}{\textbf{PSMA}}
 & DICE   & $0.507{\pm}0.022$ & $0.522{\pm}0.021$ & $0.539{\pm}0.018$ & $0.530{\pm}0.028$ & $\mathbf{0.612{\pm}0.010}$ \\
 & DMM    & $0.821{\pm}0.013$ & $0.828{\pm}0.011$ & $0.835{\pm}0.014$ & $0.829{\pm}0.011$ & $\mathbf{0.848{\pm}0.016}$ \\
 & Metric & $0.664{\pm}0.018$ & $0.675{\pm}0.016$ & $0.687{\pm}0.016$ & $0.679{\pm}0.020$ & $\mathbf{0.730{\pm}0.013}$ \\
\bottomrule
\end{tabular}
\end{table}

\paragraph{\textbf{Phase-2.}}

Due to limited computational resources, the five folds of versions Version 3 and Version 5 could not all be trained in the interactive stage. Phase~2 was therefore obtained by fine-tuning, for each tracer route, the single fold that performed best in Phase~1. The best Version 3 fold for the FDG route (trained on the combined dataset) and the best Version 5 fold for the PSMA route (trained on the PSMA-only dataset). As shown in Table~\ref{phase2_1}, the interactive scribbles improve the segmentation monotonically for both tracers, with the Dice score rising steadily from the initial prediction (w/o) through the five interaction steps (S1--S5).

The optimal scribble width differs by tracer. This is consistent with the characteristic lesion morphology of each tracer. FDG-avid lesions (melanoma, lymphoma, lung cancer) tend to be larger and more diffuse, and FDG uptake is more heterogeneous. Therefore, a wider Gaussian propagates each scribble's guidance over a larger neighbourhood, matching these lesions and providing a smoother signal that is more robust to irregular physiological uptake. PSMA lesions, in contrast, are typically small and focal with sharp, stereotyped uptake. A narrower Gaussian localizes the correction precisely on the lesion, whereas a wider one would bleed beyond its boundary into adjacent background and encourage over-segmentation.

\begin{table}[htbp]
\centering
\caption{Phase-2 results with 1-fold training and evaluation on AUTOPET V dataset}
\label{phase2_1}
\renewcommand{\arraystretch}{1.3}
\setlength{\tabcolsep}{6pt}
\begin{tabular}{cllccccccc}
\toprule
 &  &  & w/o & S1 & S2 & S3 & S4 & S5 & AUC \\
\midrule
\multirow{6}{*}{$\sigma=5$} & \multirow{3}{*}{FDG} & DICE & $0.585$ & $0.708$ & $0.803$ & $0.847$ & $0.875$ & $0.885$ & $0.784$ \\
 & & DMM & $0.681$ & $0.708$ & $0.717$ & $0.725$ & $0.734$ & $0.740$ & $0.717$ \\
 & & Metric & $0.633$ & $0.708$ & $0.760$ & $0.786$ & $0.804$ & $0.813$ & $0.751$ \\
\cmidrule{2-10}
 & \multirow{3}{*}{PSMA} & DICE & $0.581$ & $0.703$ & $0.740$ & $0.762$ & $0.778$ & $0.788$ & $\mathbf{0.725}$ \\
 & & DMM & $0.863$ & $0.869$ & $0.871$ & $0.872$ & $0.872$ & $0.874$ & $\mathbf{0.870}$ \\
 & & Metric & $0.722$ & $0.786$ & $0.806$ & $0.817$ & $0.825$ & $0.831$ & $\mathbf{0.798}$ \\
\midrule
\multirow{6}{*}{$\sigma=10$} & \multirow{3}{*}{FDG} & DICE & $0.584$ & $0.734$ & $0.811$ & $0.860$ & $0.887$ & $0.900$ & $\mathbf{0.796}$ \\
 & & DMM & $0.692$ & $0.723$ & $0.738$ & $0.746$ & $0.756$ & $0.759$ & $\mathbf{0.736}$ \\
 & & Metric & $0.638$ & $0.728$ & $0.774$ & $0.803$ & $0.822$ & $0.830$ & $\mathbf{0.766}$ \\
\cmidrule{2-10}
 & \multirow{3}{*}{PSMA} & DICE & $0.578$ & $0.692$ & $0.715$ & $0.737$ & $0.744$ & $0.753$ & $0.703$ \\
 & & DMM & $0.862$ & $0.871$ & $0.870$ & $0.871$ & $0.871$ & $0.873$ & $0.870$ \\
 & & Metric & $0.720$ & $0.781$ & $0.792$ & $0.804$ & $0.807$ & $0.813$ & $0.786$ \\
\bottomrule
\end{tabular}
\label{tab:results}
\end{table}

\section{Conclusion}
The present paper has described a two-stage, anatomy-aware and promptable segmentation system for multi-tracer PET/CT that combines organ supervision, online interactive training with simulated scribbles, and tracer-specific routing via a lightweight classifier. Organ supervision through a shared head gives the best trade-off between accuracy and robustness and was chosen as the backbone. Interactive refinement steadily improves segmentation with minimal user effort, and PSMA-specific training with a generalizable tracer classifier delivers the best tracer-wise performance while degrading gracefully on uncertain cases. 

\begin{table}[ht]
\caption{Algorithm details}\label{tab1}

\begin{tabular}{P{0.2\textwidth}P{0.2\textwidth}P{0.2\textwidth}P{0.2\textwidth}P{0.2\textwidth}} 
\toprule
\textbf{Team name} & \textbf{algorithm name} (as submitted on grand-challenge) & \textbf{data pre-processing} & \textbf{data post-processing} & \textbf{training data augmentation} \\
\midrule
UAM\_team & UAM\_team \_submission &  nnUNet Preprocessing & - & Random Brightness, Random Gamma, Random Rotation \\
\bottomrule
\end{tabular}

\vspace{2em}

\begin{tabular}{P{0.2\textwidth}P{0.2\textwidth}P{0.2\textwidth}P{0.2\textwidth}P{0.2\textwidth}} 
\toprule
\textbf{test time augmentation} & \textbf{ensembling} (e.g. cross-validation, model ensemble, ...) & \textbf{standardized framework?} (e.g. nnUNet, MONAI, ...)  & \textbf{network architecture} (e.g. UNet (3D)) & \textbf{loss} \\
\midrule
- & Tracer-based routing (combined vs PSMA-specific model); 4-fold cross-validation & nnU-Net v2 (3D) & 3D U-Net, dual encoder–decoder with shared widened lesion+organ head and guidance injection & Lesion DSC + CE + equally-weighted organ DSC + CE + scribble loss \\
\bottomrule
\end{tabular}

\vspace{2em}

\begin{tabular}{P{0.2\textwidth}P{0.2\textwidth}P{0.2\textwidth}P{0.2\textwidth}P{0.2\textwidth}} 
\toprule
\textbf{training data} & \textbf{data/model dimensionality and size} (e.g. 2D: 128x128, 3D: 128x192x160, ...) & \textbf{use of pre-trained models} (public available or own developed) & \textbf{GPU hardware for training}\\
\midrule
1014 FDG + 597 PSMA PET-CT of autoPET & Combined dataset: 128x192x160 & - & 1x Nvidia A100 \\
& PSMA dataset: 112x192x112 & & & \\
\bottomrule
\end{tabular}
\end{table}


\begin{credits}
\subsubsection{\ackname} This project has been supported by INSPIRACM (P2022/BMD7224), PowerAI+ (SI4/PJI/2024-00062 Comunidad de Madrid and UAM), Cátedra ENIA UAM-Veridas en IA Responsable (NextGenerationEU PRTR TSI-100927-2023-2), and TRUST-ID (PID2025-173396OB-I00 MICIU/AEI and the EU).

\subsubsection{\discintname}
The authors have no competing interests to declare that are
relevant to the content of this article.
\end{credits}
%
%
%
%

\end{document}